\documentclass[10pt,twocolumn,letterpaper]{article}
\usepackage[pagenumbers]{cvpr}

\usepackage[T1]{fontenc}
\usepackage{colortbl}
\usepackage{tabularx}
\usepackage{enumitem}
\usepackage[expansion=false]{microtype}
\definecolor{hlcell}{HTML}{FCEAE1}
\definecolor{wHead}{HTML}{FCE4EE}
\definecolor{wSet}{HTML}{E0ECF8}
\definecolor{wFirst}{HTML}{CA0000}
\definecolor{wSecond}{HTML}{1A9E77}
\definecolor{gook}{HTML}{1A9E77}
\definecolor{gobad}{HTML}{CA0000}
\newcommand{\grouprule}{\arrayrulecolor{black!32}\midrule\arrayrulecolor{black}}
\newcommand{\dmidrule}{\specialrule{0.4pt}{0pt}{1.1pt}\specialrule{0.4pt}{0pt}{\belowrulesep}}
\newcommand{\mh}[1]{\cellcolor{wHead}#1}
\newcommand{\sh}[1]{\cellcolor{wSet}\textbf{#1}}
\newcommand{\best}[1]{\textcolor{wFirst}{\textbf{#1}}}

\newcommand{\yes}{\textcolor{gook}{\checkmark}}
\newcommand{\no}{\textcolor{gobad}{\raisebox{.2ex}{$\scriptstyle\times$}}}

\definecolor{heatPos}{HTML}{5B9BD5}
\definecolor{heatNeg}{HTML}{E06666}

\definecolor{qOne}{HTML}{C8553D}
\definecolor{qTwo}{HTML}{D17A22}
\definecolor{qThree}{HTML}{2A9D8F}
\definecolor{qFour}{HTML}{6A4C93}
\newcommand{\qhead}[2]{\textbf{\textcolor{#1}{#2}}}
\newcommand{\qsub}[3]{\subsection{\texorpdfstring{\textcolor{#1}{#2: #3}}{#3}}}
\usepackage{caption}
\definecolor{cvprblue}{rgb}{0.13,0.40,0.67}
\usepackage[pagebackref,breaklinks,colorlinks,allcolors=cvprblue]{hyperref}

\newcommand{\diamondRecallEight}{0.932}
\newcommand{\diamondRecallZero}{0.767}

\newcommand{\freewayDmFloor}{0.0356}

\newcommand{\freewayNullEight}{0.028}

\newcommand{\freewayRecallEight}{0.153}
\newcommand{\freewayRecallZero}{0.091}

\newcommand{\freewaydmRecallZero}{0.943}

\newcommand{\nPreregHit}{4}
\newcommand{\nPreregTotal}{4}

\newcommand{\planCtrlMedian}{+0.019}
\newcommand{\planDmAcc}{0.6110}

\newcommand{\planNoneAcc}{0.5000}

\newcommand{\planRandomAcc}{0.5007}

\newcommand{\planRegretAfter}{0.141}
\newcommand{\planRegretBefore}{0.565}

\newcommand{\planRowsPos}{7}
\newcommand{\planRows}{7}

\newcommand{\pongDmFloor}{0.0278}

\newcommand{\survDeficitKeptMin}{76}

\newcommand{\freewayBudSteps}{2.8}
\newcommand{\freewayBudEv}{69}
\newcommand{\freewayBudRnd}{-4}
\newcommand{\freewayOracle}{86}
\newcommand{\pongBudSteps}{2.1}
\newcommand{\pongBudEv}{43}
\newcommand{\pongBudRnd}{5}

\newcommand{\budAbove}{8}
\newcommand{\budFams}{3}
\newcommand{\budCurves}{9}

\newcommand{\nUsable}{44}
\newcommand{\nUsablePos}{39}
\newcommand{\nUsableMed}{+44}

\newcommand{\nGamesAtari}{26}

\newcommand{\nMissed}{33}
\newcommand{\nInvented}{5}
\newcommand{\termChanMed}{0.7}

\newcommand{\nNeither}{6}

\newcommand{\gameMedNeg}{MsPacman}
\newcommand{\nDmcRows}{9}
\newcommand{\nDmcTasks}{6}

\newcommand{\dmcInLift}{+54/+54}
\newcommand{\dmcMidLift}{+8}

\newcommand{\bandMedLo}{+59}

\newcommand{\bandMedHi}{+40}

\newcommand{\winBoundMed}{1.5}

\newcommand{\topN}{44}

\newcommand{\topRnd}{17}

\newcommand{\topQ}{16}

\newcommand{\ohPrimCounted}{25}

\newcommand{\ohPrimMed}{+46}

\newcommand{\ohExtCounted}{6}

\newcommand{\ohSpearman}{0.85}

\newcommand{\pixRatioMed}{0.85}

\newcommand{\pixEvMed}{70}
\newcommand{\pixSeedSpanMed}{2.0}

\newcommand{\pxGames}{15}

\newcommand{\pxPassNames}{Freeway, CrazyClimber and Gopher}

\newcommand{\pxCounted}{8}
\newcommand{\pxEvPos}{8}
\newcommand{\pxEvMed}{+0.156}
\newcommand{\pxRndMed}{+0.012}

\newcommand{\sbFreewayDmErr}{0.171}
\newcommand{\sbFreewayMseErr}{0.096}
\newcommand{\sbFreewayDmRec}{0.94}
\newcommand{\sbFreewayMseRec}{0.95}

\newcommand{\sbPongDmRec}{0.80}
\newcommand{\sbPongMseRec}{0.69}

\newcommand{\hudCounted}{6}

\newcommand{\hudMed}{+64}

\newcommand{\sigP}{1.4\times10^{-7}}
\newcommand{\sigAbove}{38}
\newcommand{\sigBelow}{4}
\newcommand{\sigSpan}{2}

\newcommand{\sigGameP}{5.2\times10^{-4}}

\newcommand{\sensPosLo}{39}
\newcommand{\sensPosHi}{41}

\def\confName{CVPR}
\def\confYear{2027}
\title{Events, Not Pixels: Where World Models Get the Return Wrong}
\author{
  Haoyue Liu\textsuperscript{1,2}\quad
  Xiaoying Tang\textsuperscript{1,2,3,*}\\[6pt]
  {\normalfont\small
  \textsuperscript{1}School of Science and Engineering, The Chinese University of Hong Kong, Shenzhen}\\
  {\normalfont\small
  \textsuperscript{2}Shenzhen Future Network of Intelligence Institute (FNii-Shenzhen)}\\
  {\normalfont\small
  \textsuperscript{3}Guangdong Provincial Key Laboratory of Future Networks of Intelligence, CUHK(SZ)}\\[3pt]
  {\normalfont\small
  \textsuperscript{*}Corresponding author.}
}
\begin{document}
\maketitle
\raggedbottom
\begin{abstract}
World models are increasingly used for planning, where a planner ranks candidate actions by returns imagined by the model. However, these models are judged mainly by how well they render frames, even though a visually faithful rollout can still assign the wrong return. To identify where this mismatch arises, we introduce a transition-level diagnostic that replaces imagined transitions with real ones and measures the reduction in return error against a budget-matched random control. Our analysis yields two findings. (1) Event transitions, the few steps where a reward is earned or an episode ends, are rendered as well as other steps (median pixel-error ratio \pixRatioMed{}), yet restoring them removes a median of \pixEvMed{}\% of the return error. In contrast, restoring transitions with the largest MSE, SSIM or LPIPS error removes less than a random choice. Across the \nUsable{} Atari-100k model--collection pairs that pass our reliability checks, including three released families, event patches outperform random patches in \nUsablePos{} (sign test $p=\sigP{}$), with a median gain of \nUsableMed{} percentage points. (2) The remaining error is structured: a complementary patch separates errors at true events from those elsewhere, while an eight-step timing tolerance recovers few events missed by our deterministic models. This structure also guides repair: a rule recorded before four repair outcomes were known predicts when a diffusion model repairs these failures. Event patches improve plan ranking, and the same concentration appears on control tasks with rare reward events.%
\end{abstract}
\section{Introduction}
\label{sec:intro}

World models let an agent imagine the consequences of an action before taking
it~\cite{ha2018recurrent}. A planner rolls candidate action sequences forward inside the model and
acts on the ordering of their imagined
returns~\cite{chua2018deep,hafner2019learning,schrittwieser2020mastering,hansen2024td,hafner2025mastering,alonso2024diffusion,micheli2023transformers}. Released pixel world models
support this pipeline in Atari, continuous control, and
manipulation~\cite{cohen2025simulus,wu2024ivideogpt,tian2023control}, and their progress is mainly
measured by how faithfully they render the
future~\cite{wang2004image,zhang2018unreasonable,unterthiner2018towards}.

Rendering quality, however, is only loosely tied to what a planner needs. A return often hinges on
a few transitions that carry a reward or end an episode, so a model can render a rollout well yet
misprice it by missing a single paddle contact. Model-level studies expose this mismatch:
rankings by visual quality can disagree with rankings by task
success~\cite{liang2026worldlens,zhang2026world}, and whole-model oracle substitutions find that
average prediction accuracy does not consistently explain planning
performance~\cite{palenicek2023diminishing,sims2024edge,chang2026surprising}. Yet neither these
analyses nor compounding-error bounds~\cite{janner2019trust} reveal where that error occurs: is it
spread across all sixty steps of a rollout, or concentrated in only a few? We therefore ask:
\begin{center}
\emph{Where inside an imagined rollout does the error that changes a planner's decision occur?}
\end{center}

\begin{figure*}[t]
\centering
\includegraphics[width=\textwidth]{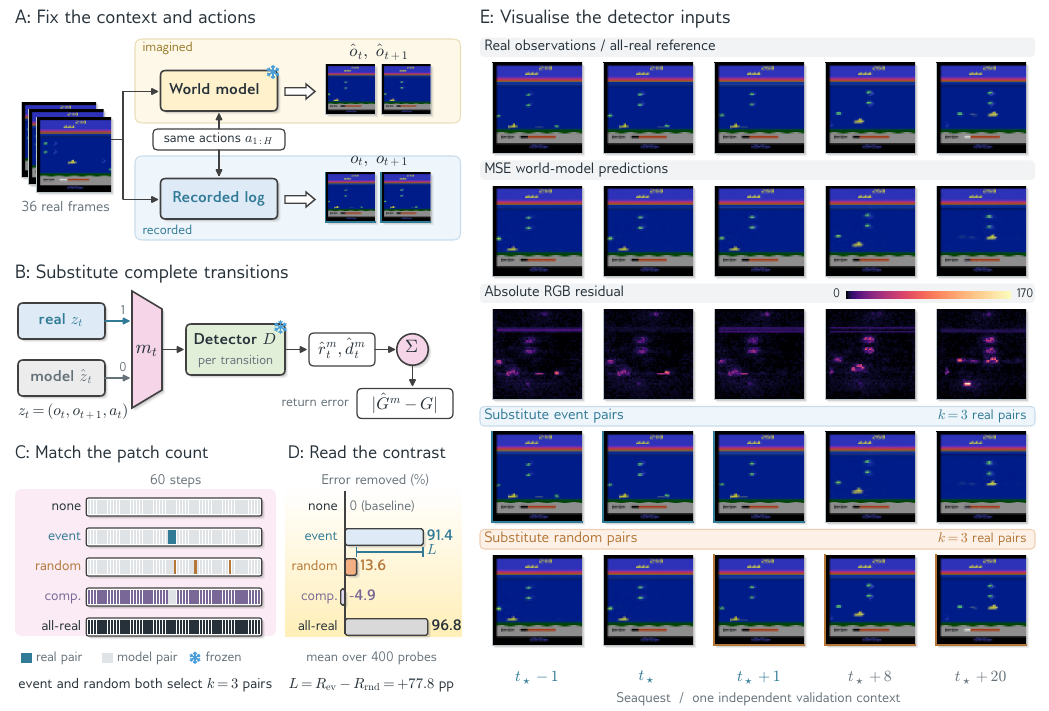}
\caption{\textbf{Return error concentrates at event transitions that pixel error does not single out.}
\textbf{A:} A saved Seaquest MSE model predicts a 60-step future from 36 real context frames
and recorded actions. \textbf{B:} Each selected detector input substitutes both frames of a
transition, keeping its action fixed. \textbf{C:} Event and random masks each select three
of sixty steps here; complement and all-real inputs complete the controls.
\textbf{D:} Saved return-error reductions over 400 probes.
\textbf{E:} Five selected times show real frames, predictions, absolute RGB residuals on a shared
square-root colour scale, and the frames supplied by the two substitution arms. Coloured
borders mark real-pair substitutions.}
\label{fig:framework}
\end{figure*}

To answer this question, we introduce a transition-level substitution diagnostic
(Fig.~\ref{fig:framework}). Each imagined rollout is paired with a real rollout that starts from
the same context and follows the same recorded actions, so the two differ only in who drew the
frames. A frozen, memoryless detector, trained once per game, reads reward and termination from
each pair of consecutive frames; replacing one imagined pair with the real one therefore changes
the return only at that step. We patch the pairs at true events, the same number of random pairs,
all non-event pairs, or every pair. These four patches isolate the targeting effect, its
budget-matched control, the error away from events, and an all-real reference.

\begin{figure*}[t]
\centering
\includegraphics[width=\textwidth]{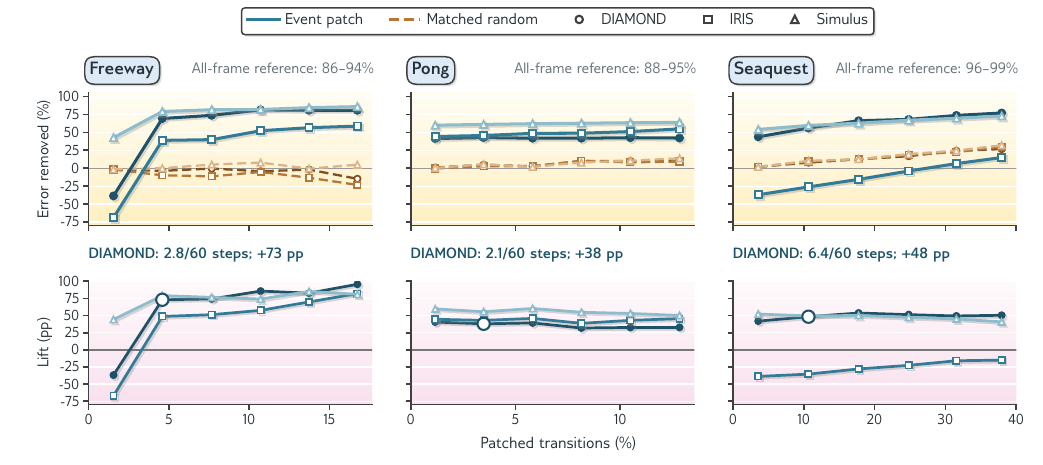}
\caption{\textbf{Targeting advantage across patch budgets.} \budCurves{} sweeps, three games,
\budFams{} released families. Solid/dashed curves denote event/random patches; markers identify
families. Lower panels show their difference, $L$, at matched coverage. Dotted guides mark
the second DIAMOND budget; annotations give its mean patched-step count and lift.
All-frame reference ranges are listed above each panel. Sweeps use one sampling seed; Table~\ref{tab:main} reports three-seed means.}

\label{fig:budget}
\end{figure*}

Our analysis yields two findings. \textbf{First, return error concentrates at events that do not
look wrong.} In Fig.~\ref{fig:framework}D, restoring three transitions around a Seaquest reward
removes almost all return error; three random ones remove little. Across \nUsable{} reliable
Atari model--collection pairs, event patches beat random patches in \nUsablePos{} by a median of
\nUsableMed{} points, and the advantage appears within a few patches (Fig.~\ref{fig:budget}).
Event frames are rendered as well as others (median pixel-error ratio \pixRatioMed{}), yet
restoring them removes a median of \pixEvMed{}\% of the return error; the same pattern holds on
eighteen games added after our predictions were recorded. \textbf{Second, the remaining error
is structured rather than merely mistimed.} Complementary patches separate errors at true events
from errors elsewhere, and timing-tolerant recall shows that temporal offset explains little
of the events our deterministic models miss. This structure also predicts where a diffusion
model helps; event patches improve plan ranking and carry over to control tasks with rare reward events.

We summarise our contributions as follows:
\begin{itemize}\setlength{\itemsep}{0pt}\setlength{\parskip}{0pt}
\item We introduce a per-transition substitution diagnostic with a budget-matched random control.
It treats the world model as a black box, and we apply it to three released model families on all \nGamesAtari{} Atari-100k games.
\item We show, for the first time on released pixel world models, that return error concentrates
at reward and termination transitions, also under each model's own reward head and without the
score display, and that a complementary patch separates errors at true events from other errors.
\item We show that a temporal offset explains little of the events missed by deterministic
models, predict all four repair outcomes with a rule recorded in advance, and show that event
patches improve plan ranking in every counted pair and carry over to control tasks with rare reward events.
\end{itemize}
\section{Related Work}
\label{sec:related}

Pixel world models support prediction and control in
games~\cite{oh2015action,chiappa2017recurrent,kaiser2019model,hafner2020mastering,robine2023transformer,zhang2023storm,micheli2023transformers,micheli2024efficient,alonso2024diffusion,cohen2025simulus}
and interactive video
simulators~\cite{yang2023learning,bruce2024genie,valevski2025diffusion,wu2024ivideogpt}. Yet visual fidelity can
diverge from physical correctness~\cite{motamed2026generative} and task
success~\cite{liang2026worldlens,zhang2026world,wang2026learning}; the same disconnect appears for
training likelihood~\cite{lambert2020objective,farahmand2017value,grimm2020value}, synthetic
training data~\cite{adamkiewicz2026pretty}, and token pruning~\cite{wang2026token}.
Concurrent work weights error by decision-relevant state dimensions~\cite{wang2026not}, while
DIAMOND shows that visual details matter~\cite{alonso2024diffusion}. We instead locate where
decision-relevant error occurs inside a rollout.

Oracle substitution localises such failures. Cloning Deterministic Worlds~\cite{xia2026cloning}
substitutes true states for learned representations, while whole-model oracle interventions relate
accuracy to planning~\cite{palenicek2023diminishing,sims2024edge,chang2026surprising}; a concurrent
diagnostic uses a kinematic null~\cite{schafer2026imagined}. Closest to ours, counterfactual
debugging~\cite{licounterfactual} assigns Shapley credit over time steps to a policy's
sim-to-real gap under injected faults, and agent explanations identify steps that decide a
return~\cite{guo2021edge,cheng2023statemask,arjona2019rudder}. We instead operate directly on
released world-model rollouts, replacing individual transition pairs under a budget-matched
random control. This brings perturbation-based
attribution~\cite{zeiler2014visualizing,fong2017interpretable,petsiuk2018rise,hooker2019benchmark},
used and tested on Atari agents~\cite{greydanus2018visualizing,atrey2019exploratory}, from pixels
to time steps.

Other work improves prediction with multi-step
training~\cite{talvitie2014model,venkatraman2015improving}, goal-aware
objectives~\cite{nair2020goal}, keyframe
targets~\cite{jayaraman2018time,pertsch2020keyframing}, region- and event-aware
objectives~\cite{lu2026improvingweakworldmodels,peng2026observations}, causal rollout structure~\cite{zhao2026transferable}, and
architecture-level probing~\cite{joseph2026interpreting}. Conditional-mean effects are also
studied in video prediction~\cite{mathieu2015deep} and perception and driving
models~\cite{porcher2026flow}. We test which explanation accounts for missed events using event
survival, a diffusion model, and the same network trained deterministically.
\section{Diagnostic Setup}
\label{sec:prelim}

\begin{figure*}[t]
\centering
\includegraphics[width=\textwidth]{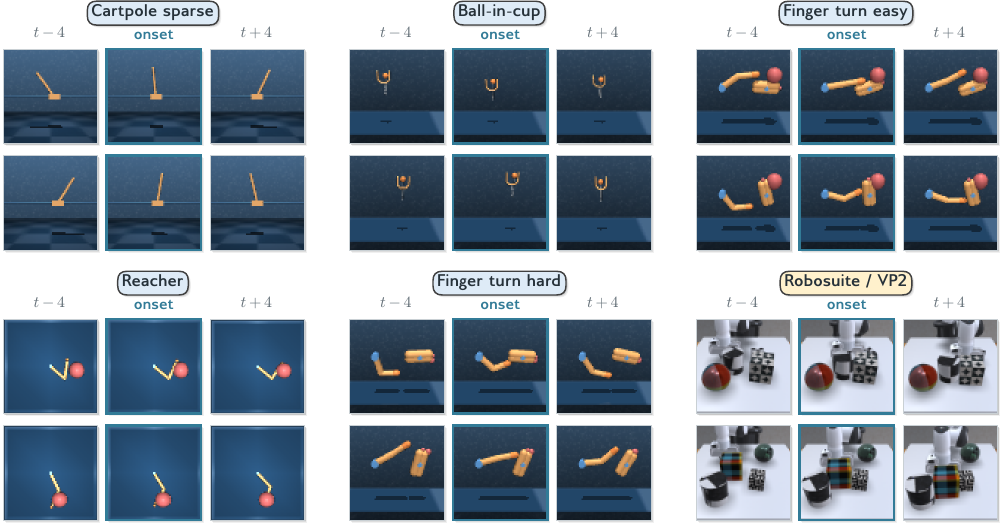}
\caption{\textbf{Task-event transitions beyond Atari.} Six tasks, two real trajectory segments per
task, and three frames per segment. Blue borders mark reward onset, the first rewarded step after at least four reward-free steps. Cases are selected
from real events, separated by more than 96 steps within each task. Aggregate targeting and
detector results are in the supplement.}

\label{fig:qual_multi}
\end{figure*}

The diagnostic needs only rendered frames, so it applies to released checkpoints without access
to their internals.

\paragraph{Return from a Predicted Trajectory.}
A \emph{probe} is a real context with fixed actions $a_{1:H}$ (Fig.~\ref{fig:framework}A); from it, a model produces imagined observations
$\hat o_{1:H}$. A frozen detector, shared by all world models of a game, reads each transition pair and its action as a reward sign and
a termination flag, $(\hat r_t,\hat d_t)=D(\hat o_t,\hat o_{t+1},a_t)$ for $t<H$; the last frame
has no successor, so $\hat r_H=\hat d_H=0$. The emulator, replaying the same actions, supplies $r_{1:H}$ and $d_{1:H}$,
and both returns stop at the first termination. Rewards enter through their sign, as in
Atari-100k training, so all games share one scale:
\begin{equation}
\hat G=\sum_{t=1}^{H}\gamma^{t-1}
\operatorname{sgn}(\hat r_t)\prod_{j<t}(1-\hat d_j),\quad
E=\mathbb{E}\lvert\hat G-G\rvert,
\label{eq:rollout}
\end{equation}
where $G$ is computed in the same way from $r$ and $d$, and $\mathbb{E}$ averages over probes.
Atari uses $H=60$ and $\gamma=0.99$.

\paragraph{Substitutions and Their Controls.}
For channel $c$ (observation $o$, reward $r$ or termination $d$) and a binary mask $m$, let $\hat G^{c,m}$ be the return after
substituting the selected real information (Fig.~\ref{fig:framework}). Observation patches replace \emph{both} frames of
the selected transition before detection, so no pair mixes real and generated frames; unselected pairs keep their predictions.
Reward and termination interventions replace $\hat r$ or $\hat d$.
We report the fractional reduction in mean absolute return error:
\begin{equation}
R(c,m)=1-\frac{\mathbb{E}[|\hat G^{c,m}-G|]}
                       {\mathbb{E}[|\hat G-G|]}.
\label{eq:attrib}
\end{equation}
The all-frame value
$R(o,\mathbf{1})$ measures performance when the detector sees only real transition pairs.

The event mask starts from nonzero real reward or termination and is dilated by one step, i.e., the
transitions just before and after are also selected, to cover delays in the visual markers. We report its realised coverage
$q=\nobreak\mathbb{E}[H^{-1}\sum_t m_t^{\rm ev}]$, which includes dilation. The random mask selects the same number of steps in each
rollout.
The targeting advantage is
\begin{equation}
L=R(o,m^{\rm ev})-R(o,m^{\rm rnd}).
\label{eq:lift}
\end{equation}
The complementary mask $\bar m^{\rm ev}$ measures errors away from the selected events, and
its contrast with the event patch classifies the failure pattern of each model.
\section{Localising the Return Error}
\label{sec:localise}

We localise the return error with one frozen, memoryless detector per game, shared by every
world model of that game. A random patch also hits true events, and more of them as coverage
grows, so the lift of Eq.~\eqref{eq:lift} credits the event patch only with what it adds beyond chance.

Figure~\ref{fig:budget} widens the event neighbourhood step by step and plots the lift against the
resulting coverage. In \budAbove{} of \budCurves{} sweeps over three games and three released
families, the event curve stays well above the random curve. For DIAMOND on Freeway, patching
\freewayBudSteps{} of sixty steps removes \freewayBudEv{}\% of the error, against
$\freewayBudRnd{}$\% for random patches and \freewayOracle{}\% when every frame is real; on Pong,
\pongBudSteps{} steps remove \pongBudEv{}\% against \pongBudRnd{}\%. The sweep also exposes a
timing effect: without dilation the Freeway event patch is negative, consistent with the reward
digit appearing after the transition that earns it, so we treat the neighbourhood as part of the
measurement.

The sweep also shows why the random control is needed. For DIAMOND and Simulus on Seaquest, both
patches remove more error as coverage grows while the lift stays flat, so the event-patch
reduction alone would overstate the targeting effect. Table~\ref{tab:main} reports every game at
a one-step neighbourhood, and Fig.~\ref{fig:qual_multi} shows events beyond Atari.

\begin{table*}[t]\centering\footnotesize
\caption{\textbf{Event targeting separates concentrated errors from diffuse failures.} Percent of return error removed by patching event (Ev), complementary (Non) or budget-matched random (Rnd) transitions (mean of three seeds); shading grows with the lift $L=\mathrm{Ev}-\mathrm{Rnd}$. Rows are sorted by event coverage $q$; grey pairs are outside the counted set, and games whose detector fails the gate are omitted (supplement).}
\label{tab:main}
\setlength{\tabcolsep}{3.2pt}
\renewcommand{\arraystretch}{1.04}
\begin{tabularx}{\textwidth}{lr*{12}{>{\raggedleft\arraybackslash}X}}
\toprule
\multicolumn{2}{l}{\sh{Game}} & \multicolumn{4}{c}{\sh{DIAMOND~\cite{alonso2024diffusion}}} & \multicolumn{4}{c}{\sh{IRIS~\cite{micheli2023transformers}}} & \multicolumn{4}{c}{\sh{Simulus~\cite{cohen2025simulus}}} \\
\cmidrule(lr){1-2}\cmidrule(lr){3-6}\cmidrule(lr){7-10}\cmidrule(lr){11-14}
\mh{} & \mh{$q$ (\%)} & \mh{Ev} & \mh{Non} & \mh{Rnd} & \mh{$L$} & \mh{Ev} & \mh{Non} & \mh{Rnd} & \mh{$L$} & \mh{Ev} & \mh{Non} & \mh{Rnd} & \mh{$L$} \\
\dmidrule
Pong & 3.5 & 41 & 46 & 5 & \cellcolor{heatPos!39}+36 & 48 & 38 & 4 & \cellcolor{heatPos!46}+45 & 63 & 23 & 4 & \cellcolor{heatPos!57}+59 \\
Kangaroo & 4.1 & 79 & 4 & 5 & \cellcolor{heatPos!68}+74 & \textcolor{gray}{98} & \textcolor{gray}{0} & \textcolor{gray}{5} & \textcolor{gray}{+92} & \textcolor{gray}{38} & \textcolor{gray}{$-$1} & \textcolor{gray}{2} & \textcolor{gray}{+36} \\
Freeway & 4.6 & 68 & $-$6 & $-$4 & \cellcolor{heatPos!66}+72 & 35 & $-$38 & $-$2 & \cellcolor{heatPos!41}+38 & 72 & $-$16 & 1 & \cellcolor{heatPos!66}+71 \\
Seaquest & 10.7 & 57 & $-$19 & 9 & \cellcolor{heatPos!48}+48 & $-$26 & 43 & 7 & \cellcolor{heatNeg!37}$-$33 & 54 & $-$28 & 8 & \cellcolor{heatPos!46}+45 \\
Frostbite & 12.0 & \textcolor{gray}{57} & \textcolor{gray}{17} & \textcolor{gray}{$-$181} & \textcolor{gray}{+238} & 71 & $-$18 & 5 & \cellcolor{heatPos!61}+65 & 83 & $-$4 & 16 & \cellcolor{heatPos!62}+66 \\
RoadRunner & 12.3 & 82 & $-$2 & 20 & \cellcolor{heatPos!59}+62 & \textcolor{gray}{91} & \textcolor{gray}{1} & \textcolor{gray}{21} & \textcolor{gray}{+70} & 81 & $-$2 & 19 & \cellcolor{heatPos!59}+62 \\
KungFuMaster & 13.9 & 93 & $-$3 & 17 & \cellcolor{heatPos!69}+75 & 83 & $-$7 & 16 & \cellcolor{heatPos!63}+67 & 92 & $-$2 & 17 & \cellcolor{heatPos!69}+75 \\
Asterix & 16.1 & 71 & 2 & 12 & \cellcolor{heatPos!57}+59 & 35 & 1 & 12 & \cellcolor{heatPos!29}+23 & 60 & $-$11 & 16 & \cellcolor{heatPos!45}+44 \\
Hero & 16.4 & 70 & 25 & 53 & \cellcolor{heatPos!26}+18 & 30 & 36 & 24 & \cellcolor{heatPos!17}+7 & 9 & 19 & 14 & \cellcolor{heatNeg!16}$-$5 \\
Boxing & 19.6 & 12 & 72 & 19 & \cellcolor{heatNeg!17}$-$7 & 79 & $-$3 & 17 & \cellcolor{heatPos!59}+62 & 70 & $-$5 & 17 & \cellcolor{heatPos!52}+53 \\
Gopher & 24.8 & 77 & $-$4 & 33 & \cellcolor{heatPos!45}+44 & 85 & $-$7 & 34 & \cellcolor{heatPos!51}+51 & 92 & $-$4 & 38 & \cellcolor{heatPos!54}+55 \\
Qbert & 25.3 & 46 & 38 & 27 & \cellcolor{heatPos!26}+19 & 92 & $-$2 & 47 & \cellcolor{heatPos!46}+45 & 77 & 9 & 42 & \cellcolor{heatPos!38}+35 \\
CrazyClimber & 38.2 & 54 & 33 & 33 & \cellcolor{heatPos!27}+20 & 58 & 26 & 34 & \cellcolor{heatPos!29}+23 & 88 & 3 & 49 & \cellcolor{heatPos!42}+40 \\
MsPacman & 42.1 & 43 & 8 & 37 & \cellcolor{heatPos!17}+7 & 30 & 69 & 33 & \cellcolor{heatNeg!14}$-$3 & 33 & 65 & 36 & \cellcolor{heatNeg!14}$-$3 \\
Krull & 58.5 & 71 & 15 & 60 & \cellcolor{heatPos!20}+11 & 90 & $-$1 & 76 & \cellcolor{heatPos!23}+14 & 93 & $-$3 & 78 & \cellcolor{heatPos!24}+16 \\
\grouprule
\multicolumn{2}{l}{\textbf{Median, counted}} & \multicolumn{3}{r}{13/14 positive} & \cellcolor{heatPos!42}\textbf{+40} & \multicolumn{3}{r}{11/13 positive} & \cellcolor{heatPos!41}\textbf{+38} & \multicolumn{3}{r}{12/14 positive} & \cellcolor{heatPos!49}\textbf{+49} \\
\midrule
\multicolumn{2}{l}{\emph{ours, matched data}} & \multicolumn{4}{c}{\emph{DIAMOND arch.}} & \multicolumn{4}{c}{\emph{RSSM}} & \multicolumn{4}{c}{\emph{Transformer}} \\
Breakout (random) & 8.3 & 13 & 35 & 0 & \cellcolor{heatPos!22}+13 & 44 & 29 & 6 & \cellcolor{heatPos!41}+38 & 92 & 1 & 6 & \cellcolor{heatPos!78}+87 \\
\bottomrule
\end{tabularx}
\end{table*}

\section{Missing Events and a Model-Class Repair}
\label{sec:repair}

\begin{figure*}[t]
\centering
\includegraphics[width=\textwidth]{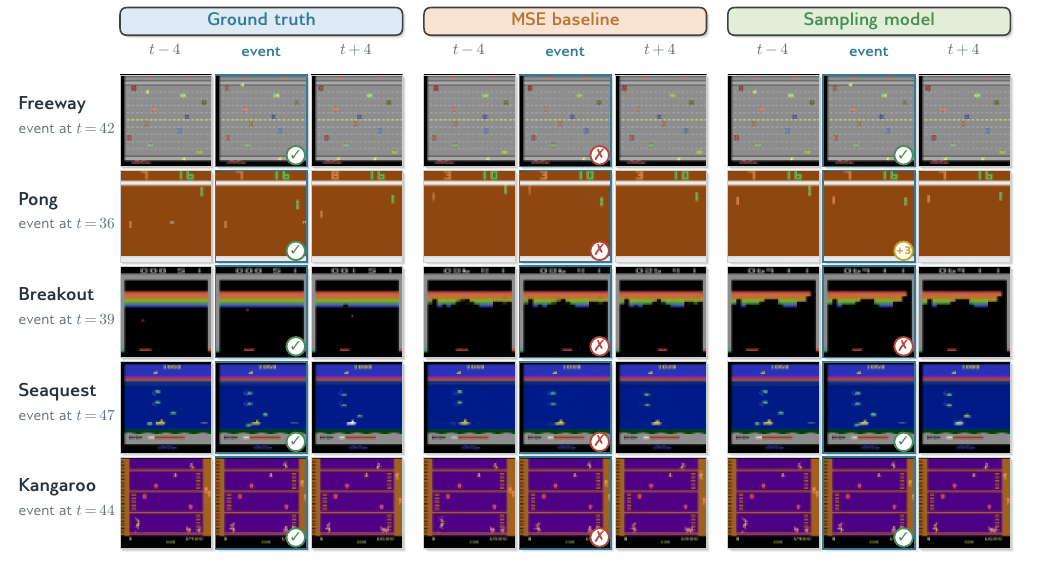}
\caption{\textbf{Testing the proposed repair across five games.}
Five games, each shown as ground truth, deterministic MSE prediction, and sampling prediction
under the same actions. Blue borders mark the real-event time; marks show whether the repair study's
detector reads the reward there (\yes), a few steps off (+3) or not (\no).
Each row uses a validation window selected from real-event timing alone.}
\label{fig:qual_overview}
\end{figure*}

Localisation shows where the return error lies, not why. We ask why imagined events are missed
and whether another model class draws them.

\subsection{Absent or Mistimed Events}
\label{sec:mean}

A detector can miss an event in imagined frames for two reasons: the model draws it at a
neighbouring step, or it does not draw it in a form the detector reads. The two predict different
curves. If events are only mistimed, recall should rise with the alignment tolerance faster than
for a displaced-prediction null; if they are not drawn readably, recall should stay low at every
tolerance.

\label{sec:classchoice}

\begin{figure*}[t]
\centering
\includegraphics[width=\textwidth]{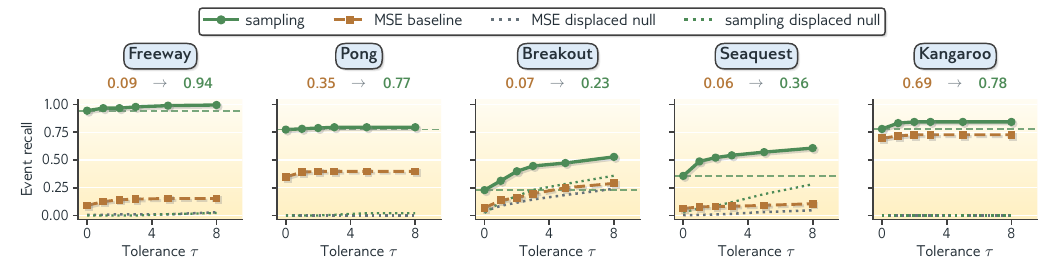}
\caption{\textbf{The event deficit of the deterministic model persists under temporal tolerance.}
Reward-event recall versus alignment tolerance across five games. Header values compare MSE
$\rightarrow$ sampling at $\tau=0$; green guides mark the sampling value.
Each displaced null matches its model's predicted event rate.
Kangaroo is the pre-specified no-effect case for return-error improvement.}
\label{fig:event_survival}
\end{figure*}

\subsection{A Rule Recorded Before the Outcomes}
\label{sec:prereg}

If events are not drawn readably, a model that draws them should reduce the return error, most
where the baseline misses most events. As a repair, we replace our deterministic MSE predictor
with a diffusion model trained on the same data~\cite{ho2020denoising,karras2022elucidating,alonso2024diffusion},
which samples each next observation instead of regressing it (``sampling'' in figures and tables).
A sampled future can differ from the realised one, so we evaluate event recall, return error, and
plan ranking.

Before observing the remaining outcomes, we pre-specified this rule: a baseline event recall at
$\tau=0$ below $0.50$ predicts a return-error reduction larger than the declared floor of that
game, and a recall at or above $0.50$ predicts no such reduction. The threshold is a heuristic for
the headroom left by missing events; floors and failure conditions were fixed at the same stage.
Freeway had already been scored, so the prospective score covers Pong and the three unrun sampling
arms. The MSE baseline gets more training, 30k updates against 8k, because at 8k updates it
detects no return on Freeway and Pong; data, sequence length, batch size, learning rate, and seeds
match (supplement).
\section{Experiments}
\label{sec:exp}

In this section, we carry out experiments to address the following questions: \qhead{qOne}{Q1:} Is the return error concentrated at
event transitions (Sec.~\ref{sec:main})? \qhead{qTwo}{Q2:} Are missed events absent or only mistimed
(Sec.~\ref{sec:survival})? \qhead{qThree}{Q3:} Does a diffusion model repair them as recorded
(Sec.~\ref{sec:results_repair})? \qhead{qFour}{Q4:} Do the same transitions drive plan selection
(Sec.~\ref{sec:planner})?

\subsection{Experimental Setup}
\label{sec:setup}

We study \nGamesAtari{} Atari games~\cite{bellemare2013arcade} with $64\times64$ observations and frameskip four, using
Atari-100k~\cite{kaiser2019model} model checkpoints~\cite{alonso2024diffusion,micheli2023transformers,cohen2025simulus}.
Across three released families and three locally trained random-policy Breakout models, \nUsable{} pairs clear reliability checks. Transfer evaluation adds \nDmcRows{} rows across \nDmcTasks{}
DeepMind Control tasks~\cite{tassa2018deepmind} and released iVideoGPT on robosuite~\cite{zhu2020robosuite,wu2024ivideogpt,tian2023control}.

\begin{figure*}[p]
\centering
\includegraphics[width=\textwidth]{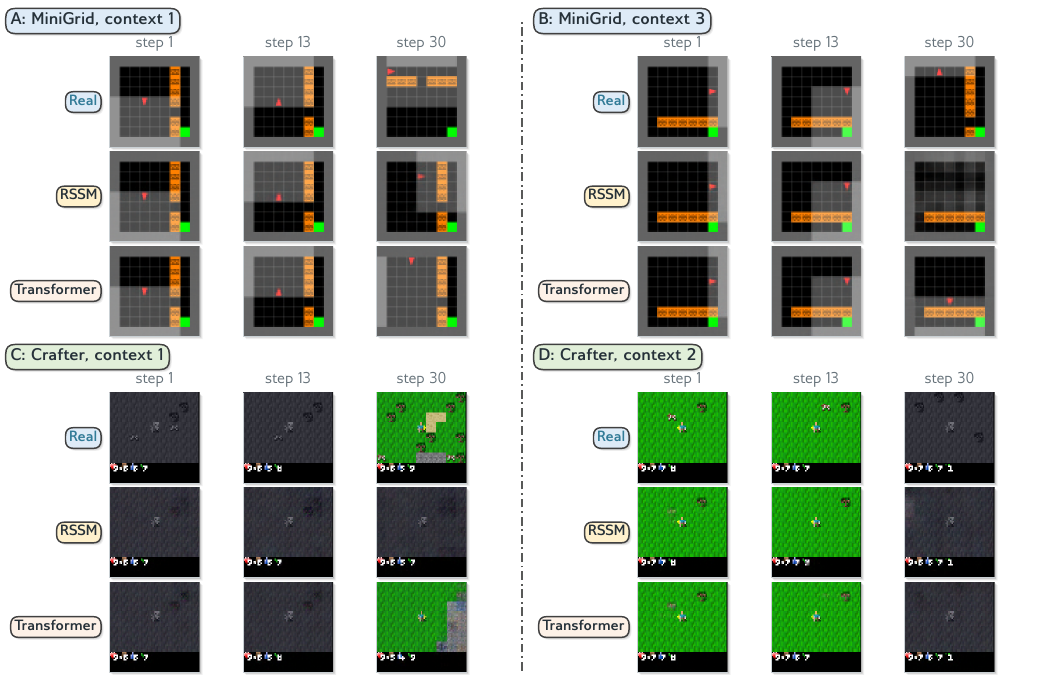}
\caption{\textbf{Scene transitions in additional environments.}
\textbf{Figures A--B} (MiniGrid) and \textbf{C--D} (Crafter) show real frames and saved RSSM/Transformer rollouts
under the same recorded actions at steps 1, 13 and 30. Map orientation and night/day
changes make trajectory divergence visible. Each rollout starts from 36 real frames;
futures may cross episode boundaries. The supplement shows additional, nonoverlapping contexts.}
\label{fig:visual_comparison}
\vspace{8pt}
\includegraphics[width=\textwidth]{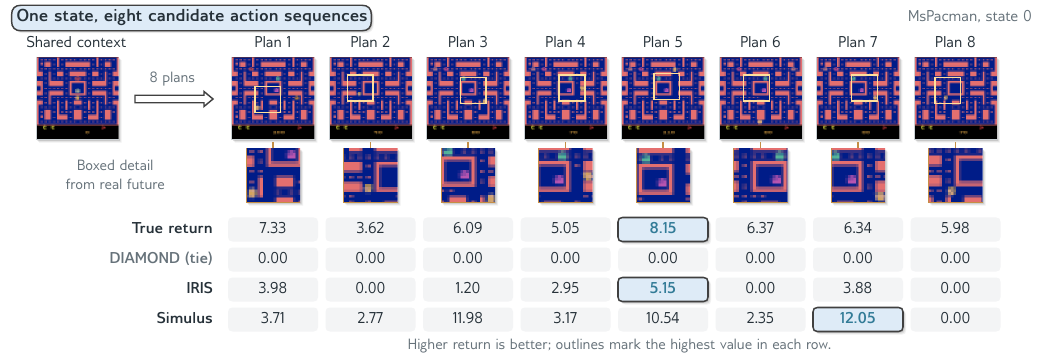}
\caption{\textbf{One MsPacman state, eight possible plans.}
The first state with nonconstant true returns shows all eight terminal-or-last real
outcomes. Insets enlarge the boxed $20\times20$ regions with greatest pixel change from
the context, excluding the HUD. Scores are unpatched; outlines mark row maxima.
DIAMOND ties all plans; IRIS selects the true-best plan 5, Simulus plan 7 (true return 6.34).
Complete planner statistics and different probe states are in the supplement.}
\label{fig:planner}
\end{figure*}

Atari attribution uses 400 probes, each paired with a real trajectory under the same actions. Each frozen detector must reach a reward-event F1 of $0.90$ on an
independent collection with environment seeds disjoint from its training data. Terminations are
read but not gated; replacing them removes a median \termChanMed{}\% of the return error. The real
reward at the last step stays in the all-real residual (median \winBoundMed{}\% of the error; supplement). Figure~\ref{fig:visual_comparison} adds matched-action
RSSM~\cite{hafner2019learning} and Transformer predictions on MiniGrid~\cite{chevalier2023minigrid} and Crafter~\cite{hafner2021benchmarking}.

\begin{table}[t]\centering\footnotesize
\caption{\textbf{Pixel error does not find the event transitions.} Every row patches as many transitions as the event patch (seed 0). A random patch removes \topRnd{}\% and hits events at rate $q$ (median \topQ{}\%).}
\label{tab:pixtopk}
\setlength{\tabcolsep}{3.2pt}
\renewcommand{\arraystretch}{1.04}
\begin{tabularx}{\columnwidth}{l*{4}{>{\raggedleft\arraybackslash}X}}
\toprule
\sh{Selection} & \multicolumn{4}{c}{\sh{Medians over the \topN{} counted pairs}} \\
\cmidrule(lr){1-1}\cmidrule(lr){2-5}
\mh{Patch chosen by} & \mh{Removed} & \mh{$L$} & \mh{$L>0$} & \mh{On events} \\
\dmidrule
Event transitions & 70\% & $+43.7$ & 39/44 & 100\% \\
\midrule
Largest MSE & 15\% & $-1.1$ & 17/44 & 19\% \\
Largest 1$-$SSIM & 11\% & $-1.2$ & 15/44 & 15\% \\
Largest LPIPS & 15\% & $-0.8$ & 17/44 & 16\% \\
\bottomrule
\end{tabularx}
\end{table}

\begin{table*}[t]\centering\footnotesize
\caption{\textbf{Selected repair comparisons.} $\Delta E=E_{\rm MSE}-E_{\rm sample}$. Errors compare competent-seed MSE means (30k updates) with designated sampling runs (8k). Freeway is the development result; shaded rows are prospective tests. \best{Red} gains exceed declared seed floors; \yes{} marks matched predictions. The complete original study, including the fourth prospective outcome, is in the supplement.}
\label{tab:repair}
\setlength{\tabcolsep}{4pt}
\renewcommand{\arraystretch}{1.18}
\begin{tabularx}{\textwidth}{ll*{7}{>{\raggedleft\arraybackslash}X}l}
\toprule
\multicolumn{2}{l}{\sh{Evaluation setting}} & \multicolumn{2}{c}{\sh{Event recall $\uparrow$}} & \multicolumn{3}{c}{\sh{Return error}} & \multicolumn{2}{c}{\sh{Decision margin}} & \sh{} \\
\cmidrule(lr){1-2}\cmidrule(lr){3-4}\cmidrule(lr){5-7}\cmidrule(lr){8-9}
\mh{Game} & \mh{Split} & \mh{MSE} & \mh{Sample} & \mh{MSE $\downarrow$} & \mh{Sample $\downarrow$} & \mh{$\Delta E\uparrow$} & \mh{Floor} & \mh{$\times$floor} & \mh{Pred. / obs.} \\
\dmidrule
Freeway & development & 0.091 & 0.943 & 0.6048 & 0.1545 & \best{+0.4503} & 0.050 & 9.0 & helps~\yes \\
\midrule
\rowcolor{hlcell}Pong & prospective & 0.348 & 0.773 & 0.2981 & 0.1421 & \best{+0.1560} & 0.079 & 2.0 & helps~\yes \\
\rowcolor{hlcell}Breakout & prospective & 0.069 & 0.229 & 1.4138 & 0.8822 & \best{+0.5316} & 0.146 & 3.6 & helps~\yes \\
\rowcolor{hlcell}Seaquest & prospective & 0.062 & 0.357 & 1.3490 & 0.5688 & \best{+0.7802} & 0.134 & 5.8 & helps~\yes \\
\bottomrule
\end{tabularx}
\end{table*}

% Two captions in one double-column float keep the visual studies on one full page.

\qsub{qOne}{A1}{Return Error Concentrates at Events}
\label{sec:main}
\label{sec:partition}

\qhead{qOne}{Main Results.} As shown in Table~\ref{tab:main}, the event patch has positive lift in
\nUsablePos{} of \nUsable{} counted pairs (sign test $p=\sigP{}$), with a median of \nUsableMed{} points (probe-bootstrap $95\%$ intervals:
\sigAbove{} above zero, \sigSpan{} spanning it, \sigBelow{} below). Each released family and every
counted game except \mbox{\gameMedNeg{}} has a positive median lift (per-game sign test,
$p=\sigGameP{}$); three sampling seeds change
lift by a median of \pixSeedSpanMed{} points, and other discounts and horizons keep
\sensPosLo{}--\sensPosHi{} pairs positive (supplement). As a patch changes only the selected transitions,
event transitions cause most of the return error.

\qhead{qOne}{Robustness to the Reward Reader.} To rule out our detector, we use each
model's reward head: all \ohPrimCounted{} pairs whose head reads real futures reliably stay positive (median
\ohPrimMed{}; rank correlation \ohSpearman{} with our detector), as do \ohExtCounted{}
pairs from games it cannot read. Hiding the score display keeps all
\hudCounted{} Pong and Freeway pairs positive (median \hudMed{}; supplement).

\qhead{qOne}{Comparison with Pixel Error.} One may expect the error to sit in poorly rendered
frames. Yet event frames are rendered as well as other frames (median ratio \pixRatioMed{}), and as
shown in Table~\ref{tab:pixtopk}, the transitions with the largest MSE, SSIM or LPIPS error are
events only about as often as random picks, so patching them removes less than random patches.
Pixel metrics thus cannot locate the return error.

\qhead{qOne}{Effect of Coverage.} Random patches hit more events as $q$ grows, so a lift that
only reflected sparse events would vanish. Yet the median lift is \bandMedHi{} above
$q=12\%$ (\bandMedLo{} below $6\%$), and all Krull pairs stay positive at $q=58.5\%$.

\qhead{qOne}{Failure Patterns.} With $R_{\rm ev}=R(o,m^{\rm ev})$ and the complementary patch's
$R_{\rm non}=R(o,\bar m^{\rm ev})$, a pair is event-dominated if
$R_{\rm ev}-R_{\rm non}>0.20$, non-event-dominated if it is below $-0.20$, and unclassified
otherwise; the counts are \nMissed{}, \nInvented{} and \nNeither{}. Simulus on
Freeway is event-dominated: event patches remove $72\%$ and the complement $-16\%$.
\mbox{DIAMOND} on Boxing is the opposite: it imagines a return of $6.67$ against a true $2.31$,
and event patches remove $12\%$ and the complement $72\%$, so under our detector its excess
return lies away from true scoring events. Under its own reward model, it is
event-dominated (supplement). Most pairs thus fail by missing true events.

\qsub{qTwo}{A2}{Missed Events Persist Under Tolerance}
\label{sec:survival}

We next ask whether missed events are merely mistimed (Sec.~\ref{sec:mean}). To investigate this, we compare reward-event recall at alignment tolerance
$\tau$ with a circularly displaced copy of the predictions, which keeps their event rate.
If events were only mistimed, widening $\tau$ would recover most of them. As shown in
Figs.~\ref{fig:qual_overview} and~\ref{fig:event_survival}, it recovers few: on Freeway, the MSE
baseline recovers \freewayRecallZero{} of the events at $\tau=0$ and \freewayRecallEight{} at
$\tau=8$, against \freewayNullEight{} for the null, and across the five games at least
\survDeficitKeptMin{}\% of its recall deficit remains at $\tau=8$. Most missed events are thus absent
rather than mistimed. \mbox{DIAMOND} on Freeway recovers more
(\diamondRecallZero{} to \diamondRecallEight{}), and the diffusion model already reaches
\freewaydmRecallZero{} at $\tau=0$.

\qsub{qThree}{A3}{A Diffusion Model Repairs as Predicted}
\label{sec:results_repair}

If missing events cause the return error, a model that draws them should reduce it. As shown in
Table~\ref{tab:repair}, all four comparisons clear their floors, and the pre-specified rule scores
\nPreregHit{}/\nPreregTotal{}, including 3/3 on the games whose detector passes the attribution
gate. Three sampling seeds on Freeway and Pong give error ranges of \freewayDmFloor{} and
\pongDmFloor{}, both below $\Delta E$.

\qhead{qThree}{Ablation on Sampling.} A natural question is whether the gain comes from sampling.
To isolate it, we train the same diffusion network with a deterministic squared-error objective
(same data, 8k updates and seeds). It still has lower error than our MSE baseline on all five
games and keeps Freeway's events (recall \sbFreewayMseRec{} against \sbFreewayDmRec{}); its
seed-mean error is lower than with sampling on Freeway (\sbFreewayMseErr{} against
\sbFreewayDmErr{}) and within the larger seed span on the other four, while sampling recalls more
events on Pong (\sbPongDmRec{} against \sbPongMseRec{}; supplement). The repair thus comes from the
diffusion network, not from sampling or the training loss.

\qsub{qFour}{A4}{Event Transitions Drive Plan Selection}
\label{sec:planner}

Finally, if events decide the imagined return, restoring them should also improve plan choice
beyond a matched random patch. To test this, we rank eight plans from each of 150 emulator states,
as a sampling-based planner does~\cite{chua2018deep,hafner2019learning}, and score pairwise accuracy
and normalised top-1 regret; \planRows{} of the twelve pairs, including the diffusion model, pass
the reliability rules (supplement), and Fig.~\ref{fig:planner} shows one. The event patch
improves pairwise accuracy in all \planRowsPos{} pairs, whereas a random patch changes it by a
median of \planCtrlMedian{}, and median top-1 regret falls from \planRegretBefore{} to
\planRegretAfter{}. Under pre-specified rules, \pxGames{}
further games add \pxCounted{} pairs from \pxPassNames{}, whose detectors also pass on
plan states, and the event patch improves all \pxEvPos{} of them (median \pxEvMed{}, random
\pxRndMed{}; supplement). On Pong, the diffusion model raises pairwise accuracy from
\planNoneAcc{} to \planDmAcc{} (random-weighted control \planRandomAcc{}). The transitions carrying the return error thus also
decide plan selection.
\section{Cross-Task Transfer and Discussion}
\label{sec:limitations}

Beyond Atari, we apply the diagnostic to six DeepMind Control tasks and robosuite, and compare
matched-action rollouts on MiniGrid and Crafter (Fig.~\ref{fig:visual_comparison}). On the control tasks of Fig.~\ref{fig:qual_multi}, the three comparisons whose detector passes yield
\dmcInLift{} points of lift on sparse cartpole across two architectures and \dmcMidLift{} on
ball-in-cup (supplement).

Concurrent work finds ball disappearance and invalid
ball--paddle interactions in five frozen Pong world models~\cite{lu2026improvingweakworldmodels}; we locate the return
error across \nGamesAtari{} games.

\section{Conclusion}
\label{sec:conclusion}
In this paper, we introduce a per-transition substitution diagnostic that locates where the
imagined return of a world model goes wrong. In \nUsablePos{} of \nUsable{} reliably measured
model--collection pairs on \mbox{Atari-100k}, the return error concentrates at reward and termination transitions that pixel metrics do not single out:
world models get the return wrong at events, in frames that look no worse than the rest.
A rule recorded in advance predicts where a diffusion model class helps, and event patches improve
plan ranking in every counted pair; the same concentration appears on control tasks
with rare reward events. These results suggest evaluating world models for planning by where their return error
occurs, in addition to how well they render frames.\label{page:body_end}
\clearpage
{\small
\bibliographystyle{ieeenat_fullname}
\bibliography{main}}
\end{document}